\documentclass[a4paper]{article}

\usepackage{INTERSPEECH2022}
\usepackage{cite}
\usepackage{graphicx}
\usepackage{pgfplots} 
\usepackage{hyperref}

\title{Low Resource Pipeline for Spoken Language Understanding via Weak Supervision}
\name{Ayush Kumar${^\dagger}$, Rishabh Kumar Tripathi$^{\dagger \star}$\thanks{$^{\dagger}$Equal contribution, $^{\star}$Work done during internship at Observe.AI}, Jithendra Vepa}
\address{
  Observe.AI, India}
\email{ayush@observe.ai, rishabh.tripathi@observe.ai, jithendra@observe.ai}

\begin{document}

\maketitle
\begin{abstract}
In Weak Supervised Learning (WSL), a model is trained over noisy labels obtained from semantic rules and task-specific pre-trained models. 
Rules offer limited generalization over tasks and require significant manual efforts while pre-trained models are available only for limited tasks.
In this work, we propose to utilize prompt-based methods as weak sources to obtain the noisy labels on unannotated data. We show that task-agnostic prompts are generalizable and can be used to obtain noisy labels for different Spoken Language Understanding (SLU) tasks such as sentiment classification, disfluency detection and emotion classification. These prompts could additionally be updated to add task-specific contexts, thus providing flexibility to design task-specific prompts. We demonstrate that prompt-based methods generate reliable labels for the above SLU tasks and thus can be used as a universal weak source to train a weak-supervised model (WSM) in absence of labeled data.
Our proposed WSL pipeline trained over prompt-based weak source outperforms other competitive low-resource benchmarks on zero and few-shot learning by more than 4\% on Macro-F1 on all of the three benchmark SLU datasets. 
The proposed method also outperforms a conventional rule based WSL pipeline by more than 5\% on Macro-F1.
\end{abstract}

\section{Introduction}
\label{section:1}
Weak supervised learning (WSL) \cite{DBLP:conf/naacl/YuZJRZZ21, DBLP:conf/emnlp/RenLSKMZ20} has been gaining interest in the research community because of the success shown by leveraging the high availability of large volumes of unlabeled data. Typically, there are three types of weak supervision \cite{zhou2018brief}: \textit{incomplete}, \textit{inexact} and \textit{inaccurate}. In \textit{incomplete} supervision, a subset of gold-labeled training data is provided and a large volume of training data remains unannotated. In \textit{inexact} supervision, training data is labeled with coarse-grained class labels instead of the exact classes. In \textit{inaccurate} supervision, no labeled training data is given but only a large volume of unannotated training data is provided, these training samples are annotated only by noisy labels. The noisy labels are derived using source(s), commonly known as weak source(s). Most common forms of weak sources observed in the area of weak supervision are: Rule-based weak sources \cite{DBLP:conf/lrec/Esuli006,DBLP:conf/icwsm/HuttoG14,DBLP:conf/msm/Nielsen11} and Task-specific pre-trained models \cite{DBLP:journals/corr/abs-2011-06993}.

Rule-based weak sources are based on the heuristics, lexicons and external knowledge bases (like SentiWordNet \cite{DBLP:conf/lrec/Esuli006}). These rules can be further bounded by a set of heuristic functions or labeling functions which maps these patterns to the class labels expected in the task. It is a challenge to extend such rules to dataset from different domains or to perform a different task. 
Additionally, the time spent in analyzing a certain dataset for creating rules is sometimes high which is often based on the complexity of the domain to which the data belongs to. However, there are a few rule-based sources already available for common tasks like sentiment, NER \textit{etc}. On the other hand, it is hard to find such readily available weak sources for tasks like disfluency detection \cite{DBLP:conf/icassp/GodfreyHM92}, emotions classification \cite{DBLP:journals/lre/BussoBLKMKCLN08} \textit{etc}. 

The other type of weak-source utilizes a pre-trained model. Such models are supposed to perform a specific task. For example, BERT-NER, which is BERT \cite{DBLP:conf/naacl/DevlinCLT19} finetuned for NER task, can be utilized as a weak source to identify named entities from the data. 
Such task-specific pre-trained models cannot be used to predict class labels for a task different than what the model is trained on.
To inject knowledge into such models for a new task requires fine-tuning the model for the new task. 
However, fine-tuning requires us to get gold annotations which demands significant manual overhead and is a time-taking task.

The common challenge imposed by both of these weak sources is the lack of generalizability across a wide variety of tasks.
Our work in this paper presents a new form of weak source which not only saves manual time and effort to write task-specific rules to generate weak labels, but can also be used to predict class labels for wide range of tasks. We present \textit{prompt-based weak source} as a potential approach to solve labeling problems across a wide range of tasks. 
\textit{Prompt-based weak source} requires prompting a Language Model (LM) \cite{DBLP:conf/acl/GaoFC20,DBLP:conf/eacl/SchickS21,DBLP:journals/corr/abs-2109-01652,logan2021cutting} to derive weak class labels. A prompt refers to a pattern string that is designed to coax the model into producing an output corresponding to a given class \cite{DBLP:conf/naacl/ScaoR21}. We study different ways to prompt LMs in \ref{section:3.1} and \ref{section:3.2}. We study the 3 key features of prompt-based sources as:
\textit{Generalizability} (We can utilize task-agnostic prompts to cater to various tasks. Effectively, we can create prompts for multiple tasks within same generic framework.),
\textit{Flexibility} (We can modify the prompts to add task and class-label specific contexts in an easy manner to improve over task-agnostic prompts.) and
\textit{Potency} (Prompting a LM has the potential to derive weak labels with reliable source performance.)
Our major contributions in this work are:
\begin{itemize}
\item We propose a generalizable and flexible low-resource setup to train weak supervised model 
using task-agnostic and task-specific prompt-based weak sources.
\item We perform extensive experiments on three benchmark SLU datasets and demonstrate the effectiveness of the proposed approach in comparison to other competitive low resource setups.
\item We release the codebase and rules developed for this work\footnote{Code will be released with camera-ready version.} for furthering the research work in this direction.
\end{itemize}


\section{Related Work}

Existing works in WSL utilize a small size of gold data during the training of a weak supervision model \cite{DBLP:conf/emnlp/WangLLYL19,DBLP:conf/emnlp/LangeHK19,DBLP:conf/acl-deeplo/HedderichK18}, while another group of work assumes that no labeled data is available during the training \cite{DBLP:conf/naacl/YuZJRZZ21,DBLP:conf/emnlp/RenLSKMZ20,DBLP:journals/vldb/RatnerBEFWR20}.  
Since we are working under a low resource constraint, we explore approaches that does not rely on labeled data to train a WSM. Ren et al. \cite{DBLP:conf/emnlp/RenLSKMZ20} 
utilized BERT to learn conditional reliability scores between multiple weak sources using an attention mechanism, while Ratner et al. \cite{DBLP:journals/vldb/RatnerBEFWR20} proposed a generative model to combine outputs from various weak sources. Yu et al. \cite{DBLP:conf/naacl/YuZJRZZ21} proposed a task-agnostic approach called COSINE which uses a contrastive self-training strategy to learn over weak labels and shows improvement over prior works in \cite{DBLP:conf/emnlp/RenLSKMZ20,DBLP:journals/vldb/RatnerBEFWR20}. 
Hence, our work utilizes COSINE \cite{DBLP:conf/naacl/YuZJRZZ21} to train a weak supervised model (WSM) considering its robustness towards high intensities of label noise.
Prompt-based methods utilize templates structured as \textit{NLI}-style prompts (section \ref{section:3.1}) or \textit{cloze}-style prompts (section \ref{section:3.2}) in a zero-shot and/or few-shot setup to predict the labels for the downstream task \cite{DBLP:conf/acl/GaoFC20,DBLP:conf/eacl/SchickS21,DBLP:journals/corr/abs-2109-01652,logan2021cutting}. Logan IV et al. \cite{logan2021cutting} demonstrated few-shot training using \textit{cloze}-style task-agnostic null-prompt.
FLAN \cite{DBLP:journals/corr/abs-2109-01652} 
utilized NLI-style instruction templates and performed instruction tuning to improve the zeroshot performance. However, due to the large size of the model (137B parameters), we find the work unsuitable to be used in creating a low resource pipeline. 
On the other hand, LMBFF \cite{DBLP:conf/acl/GaoFC20} and Pattern Exploiting Training (PET) \cite{DBLP:conf/eacl/SchickS21} utilize a relatively smaller LM (340M parameters) 
on \textit{cloze}-style prompts.
LMBFF shows that a few demonstrative examples during fine-tuning provide additional context to better learn the prompts and report improvements over PET \cite{DBLP:conf/eacl/SchickS21}. Considering the benefits of LMBFF \cite{DBLP:conf/acl/GaoFC20} over other methods in creating a low resource pipeline, we utilize this approach to perform prompt-based finetuning.

\section{Methodology}

\begin{figure}[t]
\centerline{\includegraphics[width=7cm, height=1.5cm]{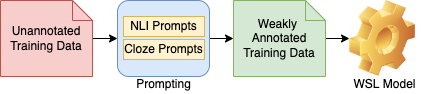}}
\caption{Prompt-based weak supervision framework}
\label{fig}
\end{figure}

The proposed methodology is a two-step process (Figure \ref{fig}). First we generate the weak labels for unlabeled data. We propose \textit{prompting the LMs} as a strategy to produce weak labels for the unlabeled training data. Next we train a weak supervised model on the previously obtained weak labels.

\subsection{Prompting as NLI}
\label{section:3.1}
In NLI-stlye prompts, the input utterance is transformed to a premise-hypothesis pair of an utterance and a prompt respectively. This transformed input is fed to an entailment model. Prompt is designed to reflect the class label of utterance if prompt (hypothesis) entails the utterance (premise). For example, for input utterance \textit{`I am happy.'} and prompt \textit{`The sentiment of the speaker is positive'},
an entailment in this case denotes that class-label is positive.
For each premise, the class label associated with the prompt having highest entailment score is treated as the weak label. For dataset with multiple classes, we frame hypothesis for each class and generate the premise-hypothesis pairs. For all such pairs of transformed inputs, the class-label associated with the pair having highest entailment score is taken as the weak label for the given utterance. For prompting, we compare a couple of pre-trained models namely \texttt{bart-large-mnli-yahoo-answers} 
and \texttt{roberta-large-mnli} 
available at Hugging Face library\footnote{https://huggingface.co/models} \cite{DBLP:journals/corr/abs-1910-03771}. 
Based on the results over the same set of prompts, we find that bart-based model predicts more accurate class labels, which we select for conducting all the NLI experiments.

\subsection{Prompting as MLM (\textit{cloze} prompts)}
\label{section:3.2}
In \textit{cloze} prompts, a piece of text is inserted in the input examples, so that the original task can be formulated as a masked language modeling problem. Inspired by LMBFF (Gao et al. \cite{DBLP:conf/acl/GaoFC20}), we leverage pre-trained \texttt{roberta-large} \cite{DBLP:journals/corr/abs-1907-11692} with an objective to fill the \texttt{[mask]} token in the prompt. 
The resulting sentence is concatenated with one demonstration per class, in the similar fashion as done by Gao et al. \cite{DBLP:conf/acl/GaoFC20}.
This scheme leverages additional context around the input sentence, making it relatively easier for LM to predict appropriate class labels. However, with the \textit{cloze} prompts, LMs may occasionally fail to understand the context properly due to the presence of noise in the data. Additionally, the failure may also be due to an attempt to solve a task beyond the capabilities of LMs. Hence, we take into account - an extended version of \textit{cloze} prompts which also requires demonstrations but performs an initial-finetuning using a few-shot setup to solve a specific task. Few-shot training only requires availability of a few gold-labeled samples. This benefits the LM to understand a specific task in low resource setting. For example, considering input utterance \textit{`I am happy.'}, the prompt based system is fed a transformed utterance containing a \texttt{[mask]} token to form utterance-instruction pair, where instruction is \textit{`The sentiment of the speaker is \texttt{[mask]}'}. For the masked token, LM generates a probability distribution over a set of verbalizers representing individual classes (Table \ref{tab:Table2}). The class corresponding to the verbalizer with highest probability is taken as the weak label.

\subsection{Weak Supervision}
Utilizing the obtained noisy (weak) labels from previous step, we leverage a weak-supervised learning architecture that learns over the weak labels by using label correction methods. 
From the results of Yu et al.\cite{DBLP:conf/naacl/YuZJRZZ21}, it is evident that COSINE framework reports better scores over other WSL methods \cite{DBLP:conf/emnlp/RenLSKMZ20,DBLP:journals/vldb/RatnerBEFWR20}.
Hence, we selectively adopt COSINE framework \cite{DBLP:conf/naacl/YuZJRZZ21} to perform WSL. COSINE leverages contrastive self-training as a strategy to empower and improve over the performance of weak source(s) iteratively. It is able to handle label noise with very high noise-ratio like the labels derived from crowd-sourcing. 
Unlike other WSMs, COSINE performs more robustly when trained upon weak labels derived from a very low coverage weak source. 

\section{Experiments}

\subsection{Dataset}
\label{section:4.1}
We consider three SLU datasets from various domains for conducting our experiments:
\textbf{CMU-MOSI} is a sentiment dataset consisting of multiple modalities  \cite{DBLP:journals/corr/ZadehZPM16}. In this work, we only consider text modality to perform sentiment classification. Similar to previous works \cite{DBLP:conf/icassp/KumarV20,DBLP:conf/acl/TsaiBLKMS19}, we binarize the class labels with train-test-valid split of 1284, 686 and 229 samples respectively.
We use the \textbf{Switchboard Disfluency} \cite{DBLP:conf/icassp/GodfreyHM92} (SWBD-D) to classify the utterances into binary class labels: \textit{fluent} and \textit{disfluent}. Train, validation and test ratio of 80:10:10 is used.
For emotion task, we utilize \textbf{IEMOCAP} \cite{DBLP:journals/lre/BussoBLKMKCLN08}. Since distinguishing between happy and excited or angry and sad is a challenging task without audio modality, in the scope of this work, we binarize the emotion in IEMOCAP dataset to positive and negative emotion.
Positive emotions comprise of happy and excited while negative emotions comprise of sad, angry and frustrated. We use the provided split of train, test and validation having 3270, 1239, 867 samples respectively.

\subsection{Various Weak Sources}
\label{section:4.2}

\textbf{Rule-based weak source}: For sentiment and emotion classification, we use SentiWordNet \cite{DBLP:conf/lrec/Esuli006} and AFINN \cite{DBLP:conf/msm/Nielsen11} lexicons along with a rule based system called VADER \cite{DBLP:conf/icwsm/HuttoG14}. We map the positive scores to positive sentiment/emotion and similarly, for negative sentiment/emotion. Empirically, we find VADER and AFINN to perform better for sentiment, while SentiWordNet and VADER perform best for emotion.
For disfluency classification on SWBD-D, the labeling functions are created based on the occurrence of filler words, repetitions and soundex \cite{odell1918soundex} codes.
We create an additional rule by aggregating the weak sources via majority voting. 
We report the average of the performance results of rule-based weak sources on Macro-F1 and Coverage metrics in Table \ref{tab:Table1}.

\begin{table}[th]
\renewcommand{\arraystretch}{1.1}
\footnotesize
\caption{Rule-based weak source performance; Coverage represents $\%$samples labeled by the weak source; Macro-F1 is reported only over the samples covered by the rules}
\label{tab:Table1}
\centering
\begin{tabular}{ |p{1.5cm}|p{1.3cm}|p{1.3cm}|  }
 \hline
 Dataset &Coverage &Macro-F1\\
 \hline
 MOSI   &74.3$\pm$4.2 &71.0$\pm$3.7\\\hline
 SWBD-D   &84.4$\pm$10.5 &73.6$\pm$7.4\\\hline
 IEMOCAP &63.9$\pm$17.2 &46.6$\pm$0.3\\\hline
\end{tabular}
\end{table}

\noindent \textbf{Prompt-based weak source}: Since rule-based weak sources have varying coverage and require significant manual efforts and time, there is a need of a low-effort method that can generate weak labels over a given dataset with reliable performance. We propose using prompting language models to obtain the weak labels. 
Specifically, we compare task-agnostic prompts and task-specific prompts over both \textit{NLI}-style and \textit{cloze}-style prompts. Task-agnostic prompt represents a general instruction that can be shared across tasks while a task-specific prompt incorporates the task information in its verbiage which helps model with an additional context.
To better understand what generalizability in prompts mean and how are the generalized (task-agnostic) prompts different from the task-specific ones, we provide representative examples of task-specific prompts in a \textit{cloze}-style setup in Table \ref{tab:Table2}, while examples of \textit{cloze}-style task-agnostic prompts are:
\begin{itemize}
\item \textit{The class best describing the text is \texttt{[mask]}}.
\item \textit{The text can be classified as \texttt{[mask]}}.
\end{itemize}
The \texttt{[mask]} is expected to be replaced by an appropriate token called \textit{verbalizer}. The \textit{verbalizer} determines the task under the consideration. By choosing different verbalizers, we can utilize same prompt for multiple tasks in a task-agnostic setting. While, in a task-specific setting, contextual prompts (Table \ref{tab:Table2}) are utilized to help predict the verbalizers more accurately. We frame the same \textit{cloze}-style prompts used in our experiments as \textit{NLI}-style prompts (i.e. hypotheses) by replacing the the \texttt{[mask]} tokens with appropriate \textit{verbalizers}. The hypotheses are entailed with the premises and yields entailment scores for each premise-hypothesis pair. The class which is a mapping to the verbalizer with highest entailment score is selected as the weak label for the premise.


\begin{table}[t]
\renewcommand{\arraystretch}{1.1}
\footnotesize
\caption{Representative examples of task-specific prompts}
\label{tab:Table2}
\centering
\begin{tabular}{ |p{1.2cm}|p{2.8cm}|p{2.7cm}|  }
 \hline
 Dataset& Task-specific prompt &\textit{verbalizer} : class label\\
 \hline
 MOSI   &The sentiment of the speaker is []. &positive : positive, negative : negative\\\hline
 SWBD-D   &The speaker [] takes a pause while speaking! &never : fluent, often : disfluent\\\hline
 IEMOCAP &I have [] emotions. &happy : positive, sad : negative\\\hline
\end{tabular}
\end{table}

\begin{table}[th]
\renewcommand{\arraystretch}{1.1}
\footnotesize
\caption{Prompt-based weak source (on Macro-F1). T-specific/agnostic denote task-specific/agnostic prompts.}
\label{tab:Table3}
\centering
\begin{tabular}{|l|c|c|c|c|}
 \hline
 Dataset& T-Specific & T-Agnostic & NLI & Cloze\\
 \hline
 MOSI&  83.5$\pm$3.3  &82.8$\pm$1.3 & 83.2$\pm$1.6 & 82.7$\pm$1.0\\\hline
 SWBD-D   &74.6$\pm$5.1    &68.9$\pm$6.5 & 42.0$\pm$15.6 & 75.2$\pm$3.9\\\hline
 IEMOCAP &68.7$\pm$3.3 &68.0$\pm$1.8 & 70.3$\pm$0.4 & 71.4$\pm$0.6 \\  \hline
\end{tabular}
\end{table}

With differences in prompts and verbalizers, the performance of the weak source vary. Thus, we choose to experiment with multiple prompts and/or verbalizers and select top prompts per task to obtain the weak labels. The mean performance of these weak sources are provided in Table \ref{tab:Table3}.
In the process, by leveraging \textit{flexibility} 
as a feature of prompt-based source, we modify the prompt structure to create various task-specific prompts. On the other other hand, the task-agnostic prompts can be utilized across various tasks performed in our experiments demonstrating \textit{generalizability} 
of the proposed approach. Additionally, 
the performance scores of prompt-based weak source reported in Table \ref{tab:Table3} shows that LMs when prompted, have potential to generate accurate weak labels (\textit{potency}).
We note that for every dataset, task-specific prompts work better than the task-agnostic prompts. This gain is significant for SWBD-D dataset denoting the need for language models to rely on task specific context unlike other tasks. Additionally, we note that NLI-style prompt produces better results for MOSI, 
while \textit{cloze}-style prompt produces more reliable labels for SWBD-D and IEMOCAP dataset (Table \ref{tab:Table3}).

\subsection{Baseline}
We compare the performance of the proposed setup with a fully-supervised and other low-resource setups like few-shot learning (FSL) and zero-shot learning (ZSL): 
\begin{itemize}
\item \textit{Oracle}: This baseline represents the performance score on test set obtained when a pre-trained RoBERTa \cite{DBLP:journals/corr/abs-1907-11692} is trained on gold labels in a fully-supervised fashion.
\item \textit{Meta-tuning (ZSL)}: This is a state-of-art work \cite{DBLP:conf/emnlp/ZhongLZK21} in ZSL where authors propose utilizing question prompts for classification tasks, where zero-shot objective is directly optimized by fine-tuning on a meta-dataset. 

\item \textit{k-Classifier (FSL)}: Since prompt fine-tuning utilize few labeled examples (16 examples), we also compare the proposed method with few-shot baselines. In this baseline, a RoBERTa model on a training set consisting of only 16 examples per class is trained.
\item \textit{DNNC (FSL)}: DNNC \cite{DBLP:conf/emnlp/ZhangHLWWYSX20}, which utilizes discriminative nearest neighbor classifier, is a state-of-art model for few-shot and out-of-scope intent prediction task. We use a 16-shot setup. 
\end{itemize}

\noindent In next set of baselines, we compare the performance of the labels obtained from different weak-sources on the test set:

\begin{itemize}
\item \textit{Rule}: We report the performance of rule-based weak sources. For the calculation of recall, the samples not covered by the rules are considered to represent false negatives.
\item \textit{Task-Agnostic-Prompt (TAP)}: Here, we report the performance score obtained by prompting the LM with a task-agnostic prompt directly over the samples in test set, without training a WSM on obtained labels.
\item \textit{Task-Specific-Prompt (TSP)}: We report the performance score obtained by prompting the LM with a task-specific prompt over the test set, without training a WSM.
\end{itemize}

\noindent Once we obtain the weak labels on the unlabeled training data, we train a weak supervised model using COSINE to obtain:
\begin{itemize}
\item \textit{C-rule}: A WSM is trained over the weak labels derived from rule-based weak source discussed in Section \ref{section:4.2}.
\end{itemize}

\noindent \textit{C-agnostic} and \textit{C-specific} are the proposed low-resource pipelines trained using COSINE on weak labels obtained in task-agnostic (\textit{TAP}) and task-specific (\textit{TSP}) prompts respectively.

\section{Results and Analysis}
We present the final results of weak-supervised learning alongside the performance of various baselines on evaluation metric of Macro-F1 in Table \ref{tab:Table4}. 
We report all of our experimental results as mean performance scores across prompts/rules along with standard deviations. For example, \textit{Rule} represents the mean of performance scores calculated using various rule-based weak sources discussed in section \ref{section:4.2} while \textit{C-Rule} represents the average scores obtained on training COSINE over various rule-based weak sources. Similarly, mean performance on prompting with various task-specific prompts is presented in \textit{TSP}, while \textit{C-specific} is the mean performance on training a WSM using \textit{TSP}. 
We also present the best scores obtained by the WSM model on each dataset (\textit{Best WSL} in Table \ref{tab:Table4}).

\begin{table}[t]
\renewcommand{\arraystretch}{1.1}
\caption{Final Results (on Macro-F1)}
\label{tab:Table4}
\centering
\begin{tabular}{l|c|c|c}
 & MOSI &SWBD-D &IEMOCAP\\
 \hline
 \textit{Oracle}    &  \textit{86.1$\pm$0.4}    &  \textit{94.5$\pm$2.0} &  \textit{80.5$\pm$0.4} \\\hline
 Meta-tuning &80.3$\pm$2.0 &49.1$\pm$2.7 &61.6$\pm$1.8 \\
 k-Classifier &73.1$\pm$7.0 &74.3$\pm$7.3 &61.4$\pm$5.9  \\
 DNNC &79.9$\pm$0.8 &63.9$\pm$1.7 &62.3$\pm$7.4 \\\hline
 Rule& 63.0$\pm$3.2 &70.9$\pm$7.1 &33.8$\pm$3.1 \\
 TAP& 81.5$\pm$1.6 &71.1$\pm$7.5 &69.0$\pm$2.8 \\
 TSP& 83.5$\pm$1.2 &73.2$\pm$5.2 &69.8$\pm$1.7 \\\hline
 C-rule&   74.6$\pm$5.4 &76.4$\pm$1.5 &41.0$\pm$1.8 \\
 C-agnostic& 82.8$\pm$1.8 &77.5$\pm$3.9 &71.5$\pm$0.3 \\
 C-specific& \textbf{84.5$\pm$0.9} & \textbf{81.9$\pm$3.7} & \textbf{71.9$\pm$0.9} \\\hline
 Best WSL &85.26 &83.86 &72.47 \\
\end{tabular}
\end{table}


\textbf{Proposed WSL vs Low-Resource Baselines}: The results show that a WSM trained on prompt based weak labels (\textit{C-agnostic, C-specific}) outperforms other baselines including state-of-the-art zero-shot (meta-tuning) and few-shot (DNNC) approaches. \textit{C-specific} outperforms the few-shot method by more than 4\% on MOSI, 7\% on SWBD-D and 9\% on IEMOCAP dataset. Further results show that training a weak supervision model (\textit{C-agnostic, C-specific}) over prompt-based weak labels bridges the gap with the Oracle model. Specifically, training a WSM improves the F1 scores by 1\% for MOSI, 8.2\% for SWBD-D and 2.1\% for IEMOCAP. This shows the effectiveness of the prompt-based weak supervised pipeline for training a low-resource model against few-shot and zero-shot methods.

\textbf{Prompt-based WSL vs Rule-based WSL}: The proposed weak supervision pipeline on prompt-based weak source also outperforms a traditional rule based weak supervision pipeline (\textit{C-rule}). The distinction between the performance of rule and prompt-based WSL pipeline is more evident for MOSI and IEMOCAP datasets. The proposed method outperforms rule based pipeline by 10\% in MOSI, 5\% in SWBD-D and 31\% in IEMOCAP dataset (\textit{C-specific} vs \textit{C-rule} in Table \ref{tab:Table4}). The higher gap in performances on MOSI and IEMOCAP could be related to the worse performance of rules, where weak labels generated from semantic rules are less accurate and have lower coverage than SWBD-D dataset. (Table \ref{tab:Table1}). We see that \textit{Rule} has particularly lower performance on IEMOCAP owing to limited coverage of such rules, while both \textit{TAP} and \textit{TSP} consistently outperforms the weak labels obtained from rules.  Thus, in addition to reducing the manual effort in writing rules for labeling the data, the prompt-based method generates more accurate labels to annotate the unlabeled data.


\textbf{Task-agnostic vs Task-specific Prompts}: We observe that weak labels obtained from task-specific prompts (\textit{TSP}) are consistently better than task-agnostic prompts (\textit{TAP}). Similar trend is observed with a WSM trained over the labels obtained from prompt-based methods, where \textit{C-specific} outperforms \textit{C-agnostic} on all tasks. While best results are obtained from task-specific prompts, even task-agnostic prompts outperform a rule based pipeline.
This demonstrates that more accurate weak labels could be obtained from a generic task-independent prompt. 
This could solve the bottleneck of labeling the data while training WSMs.

\textbf{Best WSL Scores}: Finally, we report the scores of best performing WSMs obtained on the benchmark datasets using prompt-based weak sources: 
MOSI=85.26$\%$; SWBD=83.86$\%$; IEMOCAP=72.47$\%$. We note that the best score obtained on MOSI dataset is competitive with the results on a fully supervised model (Oracle), which could be explained by a highly reliable weak labels obtained from prompt-based method on MOSI. On the other hand, the performance on SWBD-D and IEMOCAP are lower than Oracle but are strongly better than state-of-art low resource methods on zero-shot and few-shot methods. Thus, we show that low resource pipeline for SLU tasks could be effectively trained via a prompt-based weak supervision.



\section{Conclusion}
In this work, we show effectiveness of utilizing prompt-based methods as universal weak sources to develop low-resource models for wide range of benchmark SLU tasks. We show that the proposed method outperforms traditional methods of rule based weak learning as well as state-of-art methods on other low-resource settings of zero-shot and few-shot learning. In future works, we would like to study the ability of generating weak labels via automatic prompts for individual tasks and extension of work to multilingual tasks where limited training data is available.
\bibliographystyle{IEEEtran}
\bibliography{mybib}

\begin{thebibliography}{10}
\providecommand{\url}[1]{#1}
\csname url@samestyle\endcsname
\providecommand{\newblock}{\relax}
\providecommand{\bibinfo}[2]{#2}
\providecommand{\BIBentrySTDinterwordspacing}{\spaceskip=0pt\relax}
\providecommand{\BIBentryALTinterwordstretchfactor}{4}
\providecommand{\BIBentryALTinterwordspacing}{\spaceskip=\fontdimen2\font plus
\BIBentryALTinterwordstretchfactor\fontdimen3\font minus
  \fontdimen4\font\relax}
\providecommand{\BIBforeignlanguage}[2]{{%
\expandafter\ifx\csname l@#1\endcsname\relax
\typeout{** WARNING: IEEEtran.bst: No hyphenation pattern has been}%
\typeout{** loaded for the language `#1'. Using the pattern for}%
\typeout{** the default language instead.}%
\else
\language=\csname l@#1\endcsname
\fi
#2}}
\providecommand{\BIBdecl}{\relax}
\BIBdecl

\bibitem{DBLP:conf/naacl/YuZJRZZ21}
\BIBentryALTinterwordspacing
Y.~Yu, S.~Zuo, H.~Jiang, W.~Ren, T.~Zhao, and C.~Zhang, ``Fine-tuning
  pre-trained language model with weak supervision: {A} contrastive-regularized
  self-training approach,'' in \emph{Proceedings of the 2021 Conference of the
  North American Chapter of the Association for Computational Linguistics:
  Human Language Technologies, {NAACL-HLT} 2021}, 2021, pp. 1063--1077.
  [Online]. Available: \url{https://doi.org/10.18653/v1/2021.naacl-main.84}
\BIBentrySTDinterwordspacing

\bibitem{DBLP:conf/emnlp/RenLSKMZ20}
\BIBentryALTinterwordspacing
W.~Ren, Y.~Li, H.~Su, D.~Kartchner, C.~Mitchell, and C.~Zhang, ``Denoising
  multi-source weak supervision for neural text classification,'' in
  \emph{Findings of the Association for Computational Linguistics: {EMNLP}
  2020, Online Event}, ser. Findings of {ACL}, vol. {EMNLP} 2020, 2020, pp.
  3739--3754. [Online]. Available:
  \url{https://doi.org/10.18653/v1/2020.findings-emnlp.334}
\BIBentrySTDinterwordspacing

\bibitem{zhou2018brief}
Z.-H. Zhou, ``A brief introduction to weakly supervised learning,''
  \emph{National science review}, vol.~5, no.~1, pp. 44--53, 2018.

\bibitem{DBLP:conf/lrec/Esuli006}
\BIBentryALTinterwordspacing
A.~Esuli and F.~Sebastiani, ``{SENTIWORDNET:} {A} publicly available lexical
  resource for opinion mining,'' in \emph{Proceedings of the Fifth
  International Conference on Language Resources and Evaluation, {LREC}, Genoa,
  Italy}, 2006, pp. 417--422. [Online]. Available:
  \url{http://www.lrec-conf.org/proceedings/lrec2006/summaries/384.html}
\BIBentrySTDinterwordspacing

\bibitem{DBLP:conf/icwsm/HuttoG14}
\BIBentryALTinterwordspacing
C.~J. Hutto and E.~Gilbert, ``{VADER:} {A} parsimonious rule-based model for
  sentiment analysis of social media text,'' in \emph{Proceedings of the Eighth
  International Conference on Weblogs and Social Media, {ICWSM}, Michigan,
  USA}, 2014. [Online]. Available:
  \url{http://www.aaai.org/ocs/index.php/ICWSM/ICWSM14/paper/view/8109}
\BIBentrySTDinterwordspacing

\bibitem{DBLP:conf/msm/Nielsen11}
\BIBentryALTinterwordspacing
F.~{\AA}. Nielsen, ``A new {ANEW:} evaluation of a word list for sentiment
  analysis in microblogs,'' in \emph{Proceedings of the {ESWC2011} Workshop on
  'Making Sense of Microposts': Big things come in small packages, Crete,
  Greece}, ser. {CEUR} Workshop Proceedings, vol. 718, 2011, pp. 93--98.
  [Online]. Available: \url{http://ceur-ws.org/Vol-718/paper\_16.pdf}
\BIBentrySTDinterwordspacing

\bibitem{DBLP:journals/corr/abs-2011-06993}
\BIBentryALTinterwordspacing
S.~Schweter and A.~Akbik, ``{FLERT:} document-level features for named entity
  recognition,'' \emph{CoRR}, vol. abs/2011.06993, 2020. [Online]. Available:
  \url{https://arxiv.org/abs/2011.06993}
\BIBentrySTDinterwordspacing

\bibitem{DBLP:conf/icassp/GodfreyHM92}
\BIBentryALTinterwordspacing
J.~J. Godfrey, E.~Holliman, and J.~McDaniel, ``{SWITCHBOARD:} telephone speech
  corpus for research and development,'' in \emph{1992 {IEEE} International
  Conference on Acoustics, Speech, and Signal Processing, {ICASSP}, California,
  USA}, 1992, pp. 517--520. [Online]. Available:
  \url{https://doi.org/10.1109/ICASSP.1992.225858}
\BIBentrySTDinterwordspacing

\bibitem{DBLP:journals/lre/BussoBLKMKCLN08}
\BIBentryALTinterwordspacing
C.~Busso \emph{et~al.}, ``{IEMOCAP:} interactive emotional dyadic motion
  capture database,'' \emph{Lang. Resour. Evaluation}, vol.~42, no.~4, pp.
  335--359, 2008. [Online]. Available:
  \url{https://doi.org/10.1007/s10579-008-9076-6}
\BIBentrySTDinterwordspacing

\bibitem{DBLP:conf/naacl/DevlinCLT19}
\BIBentryALTinterwordspacing
J.~Devlin, M.~Chang, K.~Lee, and K.~Toutanova, ``{BERT:} pre-training of deep
  bidirectional transformers for language understanding,'' in \emph{Proceedings
  of the 2019 Conference of the North American Chapter of the Association for
  Computational Linguistics: Human Language Technologies, {NAACL-HLT} 2019,
  Minneapolis, USA}, 2019, pp. 4171--4186. [Online]. Available:
  \url{https://doi.org/10.18653/v1/n19-1423}
\BIBentrySTDinterwordspacing

\bibitem{DBLP:conf/acl/GaoFC20}
\BIBentryALTinterwordspacing
T.~Gao, A.~Fisch, and D.~Chen, ``Making pre-trained language models better
  few-shot learners,'' in \emph{Proceedings of the 59th Annual Meeting of the
  Association for Computational Linguistics and the 11th International Joint
  Conference on Natural Language Processing, {ACL/IJCNLP} 2021}, 2021, pp.
  3816--3830. [Online]. Available:
  \url{https://doi.org/10.18653/v1/2021.acl-long.295}
\BIBentrySTDinterwordspacing

\bibitem{DBLP:conf/eacl/SchickS21}
\BIBentryALTinterwordspacing
T.~Schick and H.~Sch{\"{u}}tze, ``Exploiting cloze-questions for few-shot text
  classification and natural language inference,'' in \emph{Proceedings of the
  16th Conference of the European Chapter of the Association for Computational
  Linguistics: Main Volume, {EACL}}, 2021, pp. 255--269. [Online]. Available:
  \url{https://doi.org/10.18653/v1/2021.eacl-main.20}
\BIBentrySTDinterwordspacing

\bibitem{DBLP:journals/corr/abs-2109-01652}
\BIBentryALTinterwordspacing
J.~Wei \emph{et~al.}, ``Finetuned language models are zero-shot learners,''
  \emph{CoRR}, vol. abs/2109.01652, 2021. [Online]. Available:
  \url{https://arxiv.org/abs/2109.01652}
\BIBentrySTDinterwordspacing

\bibitem{logan2021cutting}
\BIBentryALTinterwordspacing
R.~L. Logan~IV, I.~Bala{\v{z}}evi{\'c}, E.~Wallace, F.~Petroni, S.~Singh, and
  S.~Riedel, ``Cutting down on prompts and parameters: Simple few-shot learning
  with language models,'' \emph{arXiv preprint arXiv:2106.13353}, 2021.
  [Online]. Available: \url{https://doi.org/10.48550/arXiv.2106.13353}
\BIBentrySTDinterwordspacing

\bibitem{DBLP:conf/naacl/ScaoR21}
\BIBentryALTinterwordspacing
T.~L. Scao and A.~M. Rush, ``How many data points is a prompt worth?'' in
  \emph{NAACL-HLT, 2021, Online}, 2021, pp. 2627--2636. [Online]. Available:
  \url{https://doi.org/10.18653/v1/2021.naacl-main.208}
\BIBentrySTDinterwordspacing

\bibitem{DBLP:conf/emnlp/WangLLYL19}
\BIBentryALTinterwordspacing
H.~Wang, B.~Liu, C.~Li, Y.~Yang, and T.~Li, ``Learning with noisy labels for
  sentence-level sentiment classification,'' in \emph{Proceedings of the 2019
  Conference on Empirical Methods in Natural Language Processing and the 9th
  International Joint Conference on Natural Language Processing, {EMNLP-IJCNLP}
  2019, Hong Kong}, 2019, pp. 6285--6291. [Online]. Available:
  \url{https://doi.org/10.18653/v1/D19-1655}
\BIBentrySTDinterwordspacing

\bibitem{DBLP:conf/emnlp/LangeHK19}
\BIBentryALTinterwordspacing
L.~Lange, M.~A. Hedderich, and D.~Klakow, ``Feature-dependent confusion
  matrices for low-resource {NER} labeling with noisy labels,'' in
  \emph{Proceedings of the 2019 Conference on Empirical Methods in Natural
  Language Processing and the 9th International Joint Conference on Natural
  Language Processing, {EMNLP-IJCNLP} 2019, Hong Kong}, 2019, pp. 3552--3557.
  [Online]. Available: \url{https://doi.org/10.18653/v1/D19-1362}
\BIBentrySTDinterwordspacing

\bibitem{DBLP:conf/acl-deeplo/HedderichK18}
\BIBentryALTinterwordspacing
M.~A. Hedderich and D.~Klakow, ``Training a neural network in a low-resource
  setting on automatically annotated noisy data,'' in \emph{Proceedings of the
  Workshop on Deep Learning Approaches for Low-Resource NLP, DeepLo@ACL 2018,
  Melbourne, Australia}, 2018, pp. 12--18. [Online]. Available:
  \url{https://doi.org/10.18653/v1/W18-3402}
\BIBentrySTDinterwordspacing

\bibitem{DBLP:journals/vldb/RatnerBEFWR20}
\BIBentryALTinterwordspacing
A.~Ratner, S.~H. Bach, H.~R. Ehrenberg, J.~A. Fries, S.~Wu, and C.~R{\'{e}},
  ``Snorkel: rapid training data creation with weak supervision,'' \emph{{VLDB}
  J.}, vol.~29, no. 2-3, pp. 709--730, 2020. [Online]. Available:
  \url{https://doi.org/10.1007/s00778-019-00552-1}
\BIBentrySTDinterwordspacing

\bibitem{DBLP:journals/corr/abs-1910-03771}
\BIBentryALTinterwordspacing
T.~Wolf \emph{et~al.}, ``Huggingface's transformers: State-of-the-art natural
  language processing,'' \emph{CoRR}, vol. abs/1910.03771, 2019. [Online].
  Available: \url{http://arxiv.org/abs/1910.03771}
\BIBentrySTDinterwordspacing

\bibitem{DBLP:journals/corr/abs-1907-11692}
\BIBentryALTinterwordspacing
Y.~Liu \emph{et~al.}, ``Roberta: {A} robustly optimized {BERT} pretraining
  approach,'' \emph{CoRR}, vol. abs/1907.11692, 2019. [Online]. Available:
  \url{http://arxiv.org/abs/1907.11692}
\BIBentrySTDinterwordspacing

\bibitem{DBLP:journals/corr/ZadehZPM16}
\BIBentryALTinterwordspacing
A.~Zadeh, R.~Zellers, E.~Pincus, and L.~Morency, ``{MOSI:} multimodal corpus of
  sentiment intensity and subjectivity analysis in online opinion videos,''
  \emph{CoRR}, vol. abs/1606.06259, 2016. [Online]. Available:
  \url{http://arxiv.org/abs/1606.06259}
\BIBentrySTDinterwordspacing

\bibitem{DBLP:conf/icassp/KumarV20}
\BIBentryALTinterwordspacing
A.~Kumar and J.~Vepa, ``Gated mechanism for attention based multi modal
  sentiment analysis,'' in \emph{2020 {IEEE} International Conference on
  Acoustics, Speech and Signal Processing, {ICASSP} 2020, Barcelona, Spain},
  2020, pp. 4477--4481. [Online]. Available:
  \url{https://doi.org/10.1109/ICASSP40776.2020.9053012}
\BIBentrySTDinterwordspacing

\bibitem{DBLP:conf/acl/TsaiBLKMS19}
\BIBentryALTinterwordspacing
Y.~H. Tsai, S.~Bai, P.~P. Liang, J.~Z. Kolter, L.~Morency, and
  R.~Salakhutdinov, ``Multimodal transformer for unaligned multimodal language
  sequences,'' in \emph{Proceedings of the 57th Conference of the Association
  for Computational Linguistics, {ACL} 2019, Florence, Italy}, 2019, pp.
  6558--6569. [Online]. Available: \url{https://doi.org/10.18653/v1/p19-1656}
\BIBentrySTDinterwordspacing

\bibitem{odell1918soundex}
K.~Odell and R.~Russell, ``Soundex phonetic comparison system,'' \emph{US
  Patent}, vol. 1261167, 1918.

\bibitem{DBLP:conf/emnlp/ZhongLZK21}
\BIBentryALTinterwordspacing
R.~Zhong, K.~Lee, Z.~Zhang, and D.~Klein, ``Adapting language models for
  zero-shot learning by meta-tuning on dataset and prompt collections,'' in
  \emph{Findings of the Association for Computational Linguistics: {EMNLP}
  2021, Virtual Event / Punta Cana, Dominican Republic}, 2021, pp. 2856--2878.
  [Online]. Available:
  \url{https://doi.org/10.18653/v1/2021.findings-emnlp.244}
\BIBentrySTDinterwordspacing

\bibitem{DBLP:conf/emnlp/ZhangHLWWYSX20}
\BIBentryALTinterwordspacing
J.~Zhang \emph{et~al.}, ``Discriminative nearest neighbor few-shot intent
  detection by transferring natural language inference,'' in \emph{Proceedings
  of the 2020 Conference on Empirical Methods in Natural Language Processing,
  {EMNLP} 2020, Online}, 2020, pp. 5064--5082. [Online]. Available:
  \url{https://doi.org/10.18653/v1/2020.emnlp-main.411}
\BIBentrySTDinterwordspacing

\end{thebibliography}
\end{document}